%% file: main.tex
\documentclass[10pt,twocolumn,letterpaper]{article}

\usepackage[pagenumbers]{cvpr} 

\usepackage{times}
\usepackage{epsfig}
\usepackage{graphicx}
\usepackage{amsmath}
\usepackage{amssymb}
\usepackage{subcaption}
\usepackage{blindtext}
\usepackage{booktabs}  
\usepackage{multirow}
\usepackage{enumitem}
\usepackage{pifont}
\usepackage{tabulary,multirow,overpic,xcolor,subfloat}
\usepackage{makecell}
\usepackage{colortbl}
\usepackage{mathtools}
\usepackage[pagebackref=true, breaklinks=true, letterpaper=true, colorlinks, bookmarks=false]{hyperref}
\definecolor{citecolor}{HTML}{023fba}
\definecolor{graycolor}{rgb}{0.95,0.95,0.95}

\usepackage[capitalize]{cleveref}
\crefname{section}{Sec.}{Secs.}
\Crefname{section}{Section}{Sections}
\Crefname{table}{Table}{Tables}
\crefname{table}{Tab.}{Tabs.}

\newcommand{\R}{\mathbb{R}}

\def\cvprPaperID{} 
\def\confName{CVPR}
\def\confYear{}

\begin{document}
\title{Mixed Attention Network for Hyperspectral Image Denoising}

\author{Zeqiang Lai ~ Ying Fu\\
Beijing Institute of Technology\\
{\tt\small \{laizeqiang, fuying\}@bit.edu.cn}
}

\maketitle

\input{sections.tex}

 {\small
  \bibliographystyle{ieee_fullname}
  \bibliography{egbib}
 }

\end{document}

%% file: sections.tex
\begin{abstract}
   Hyperspectral image denoising is unique for the highly similar and correlated spectral information that should be properly considered. However, existing methods show limitations in exploring the spectral correlations across different bands and feature interactions within each band. Besides, the low- and high-level features usually exhibit different importance for different spatial-spectral regions, which is not fully explored for current algorithms as well.
   In this paper, we present a Mixed Attention Network (MAN) that simultaneously considers the inter- and intra-spectral correlations as well as the interactions between low- and high-level spatial-spectral meaningful features. Specifically, we introduce a multi-head recurrent spectral attention that efficiently integrates the inter-spectral features across all the spectral bands. These features are further enhanced with a progressive spectral channel attention by exploring the intra-spectral relationships. Moreover, we propose an attentive skip-connection that adaptively controls the proportion of the low- and high-level spatial-spectral features from the encoder and decoder to better enhance the aggregated features.
   Extensive experiments show that our MAN outperforms existing state-of-the-art methods on simulated and real noise settings while maintaining a low cost of parameters and running time.
     Code is available at \url{https://github.com/Zeqiang-Lai/MAN}.
\end{abstract}

\section{Introduction}

Hyperspectral image (HSI) is made up of numerous bands across a wide range of the spectrum. Different from common color images, HSIs divide the spectrum into much more bands than three and their wavelengths can be extended beyond the visible. Such properties make HSI especially attractive and useful for the applications of remote sensing \cite{blackburn2007hyperspectral,thenkabail2016hyperspectral,zhang2010object}, face recognition \cite{1251148,zhao2014sparse}, classification \cite{azar2020hyperspectral,zhou2019hyperspectral,cao2019hyperspectral}, \etc.
However, due to the limitation of existing hyperspectral imaging techniques, the captured HSIs often suffer from severe corruption, which makes the development of robust denoising algorithms an urgent need.

\begin{figure}[t]
   \centering
   \includegraphics[width=1\linewidth]{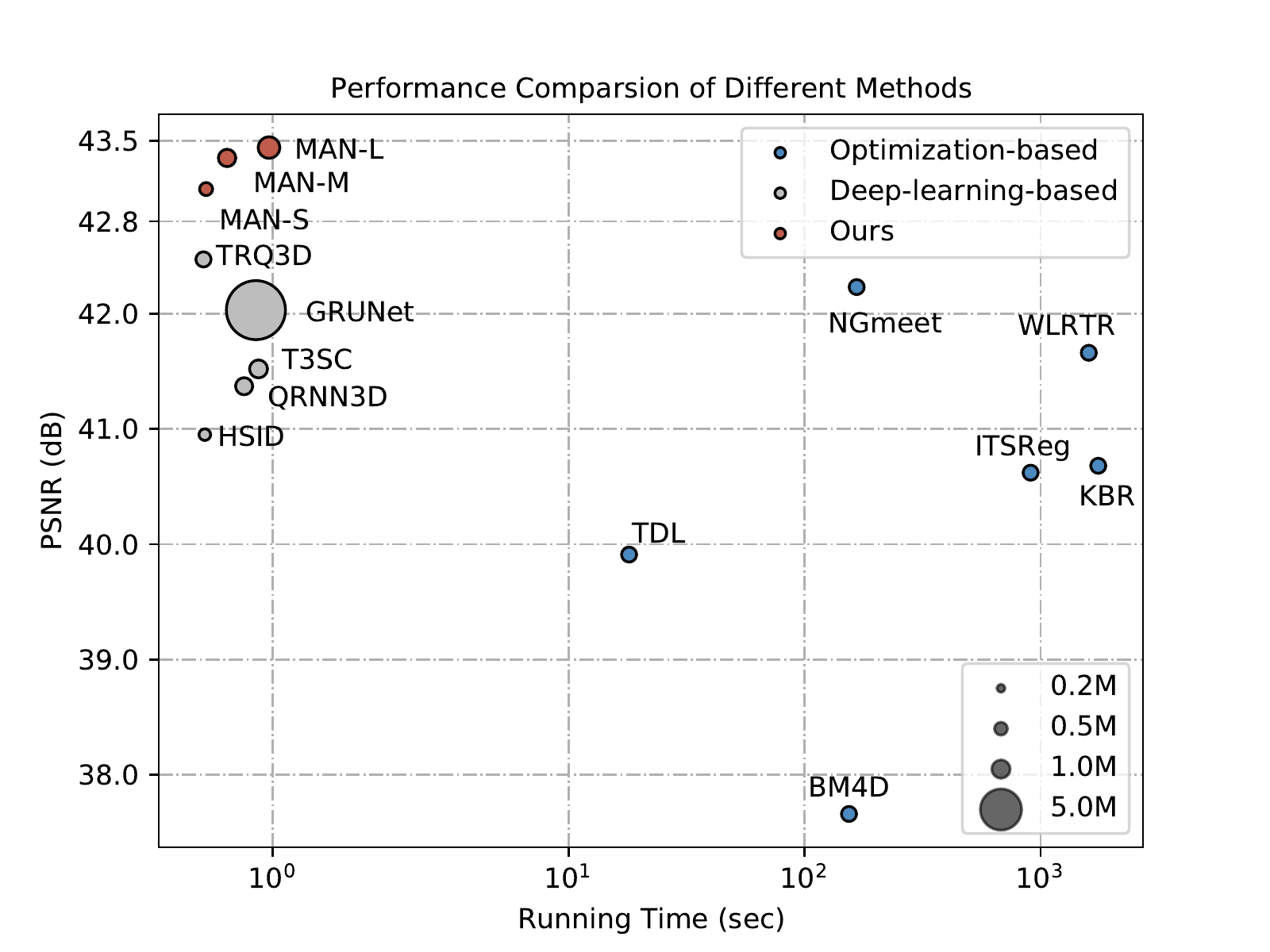}
   \caption{Speed and performance comparison. Our method outperforms state-of-the-art methods with low-cost parameters and runtime. Comparisons are performed on the ICVL dataset with blind Gaussian noise.}
   \label{fig:speed-performance}
\end{figure}

Traditionally, optimization algorithms are often adopted to solve HSI denoising with different hand-crafted priors exploring the domain knowledge of the HSI, \ie, global correlation along the spectrum and spatial-spectral correlation. Typical priors that are extensively studied includes, total variation \cite{yuan2012hyperspectral,wang2017hyperspectral}, wavelet \cite{othman2006noise}, low-rank  \cite{zhao2020fast,sun2017hyperspectral,wei2019low}, and etc. By considering the spectral and spatial redundancy, non-local patch-similarity \cite{maggioni2012nonlocal} is also widely used in conjugation with variable splitting algorithms \cite{he2019non} and tensor-based dictionary learning \cite{peng2014decomposable}. These methods generally have no requirement for a large amount of training data or even be training-free, but most of them are time-consuming (see Figure \ref{fig:speed-performance}) and their performance is strongly correlated with the matching degree of handcrafted priors with the underlying characteristics of HSIs, which weakens their performance for real-world denoising with complex noise.

Recent works on HSI denoising shift their attention to the learning-based approaches to model complex intrinsic characteristics of HSI, \eg, inter-spectral correlations with sliding-window convolution \cite{yuan2018hyperspectral}, and local spatial-spectral correlations with 3D convolution \cite{wei20203}. Augmented with various commonly used techniques, \eg, residual connection \cite{yuan2018hyperspectral}, skip connection \cite{dong2019deep}, and recurrent network \cite{wei20203}, these methods have obtained much better results than optimization-based ones. Nevertheless, their performance and efficiency are still limited and the thorough consideration of inter- and intra-spectral correlations is still lacking.
Besides, though most HSI restoration models \cite{dong2019deep,wei20203} employ skip connections to prevent the loss of low-level information, the development of such modules mostly stays at the original additive or concat versions. However, these vanilla ones either neglect the different importance of the features from encoder and decoder, or mix the tasks of features weighting and processing, which makes them less effective for handling diverse information from different feature channels and spectral bands.
Therefore, it is foreseeable that whether a model can comprehensively take information from different aspects into account would be the key to further boosting the HSI denoising performance.

In this paper, we propose a Mixed Attention Network (MAN) for hyperspectral image denoising, which simultaneously considers the inter- and intra-spectral correlations as well as the interactions between low- and high-level spatial-spectral meaningful features. Specifically, we introduce a Multi-Head Recurrent Spectral Attention (MHRSA) block in place of vanilla spectral fusion methods \cite{wei20203,yuan2018hyperspectral}. It applies recurrent attention across the spectral dimension to dynamically mix the inter-spectral features from all spectral bands. MHRSA adopts two simple MLPs rather than dense convolutions to directly transform the input features into candidate values and attention maps, which makes it not only computationally efficient but also lightweight in terms of model parameters. Another important characteristic of our MHRSA is the efficient bi-directional context aggregation through a multi-head feature partition. This is achieved by splitting rather than repeating the features into two heads and performing parallel recurrent attention in reverse directions. Apart from inter-spectral correlations, we propose a Progressive Spectral Channel Attention (PSCA) that sequentially applies the dynamic channel mixing and the static channel mixing to progressively mix the intra-spectral features for each band. The MHRSA and PSCA cooperate with each other and enable the informative features to propagate and interact along both inter-spectral and intra-spectral dimensions, which results in exceedingly discriminative features for subsequent reconstruction.

Furthermore, we reexamine the skip connections of the U-shaped models \cite{dong2019deep}, which are used to ensure the preservation of fine-grained details. By analyzing the underlying operations of the most commonly used additive and concat versions, we propose a novel Attentive Skip Connection (ASC) that adaptively combines the low- and high-level features from encoder and decoder. The ASC assigns element-wise fusion weight for each spatial-spectral location, which allows subsequent layers to selectively focus on more informative regions, thus leading to better denoising results. We conduct comprehensive experiments and demonstrate the state-of-the-art performance of our MAN on various HSI denoising datasets under simulated and real-world noise.

Our contributions are summarized as follows:
\begin{itemize}[noitemsep,topsep=2pt,leftmargin=*]
   \item We propose a mixed attention network for hyperspectral image denoising, which simultaneously considers the inter- and intra-spectral as well as low- and high-level spatial-spectral feature correlations.
   \item We introduce a multi-head spectral recurrent attention block to dynamically aggregate the inter-spectral features, and a progressive spectral channel attention block for integrating the intra-spectral features.
   \item We present an attentive skip connection to adaptively strengthen the important spatial-spectral meaningful features that flow forward from low- to high-level.
\end{itemize}

\section{Related Works}

The methods for HSI denoising could generally be divided into optimization-based methods, deep-learning methods, as well as hybrid methods. In this section, we provide an overview of their recent major approaches.

\subsection{Optimization-based Methods}

Traditional methods solve the HSI denoising by treating it as an optimization problem, where they attempt to find unknown clean HSI by minimizing an optimization objective that incorporates the properties of spectrum and images. Such incorporation is generally achieved by designing different hand-crafted priors, e.g., total variation priors \cite{yuan2012hyperspectral,wang2017hyperspectral}, wavelet priors \cite{othman2006noise}, and low-rank priors \cite{zhao2020fast,sun2017hyperspectral,wang2017hyperspectral,wei2019low}. By considering the non-local self-similarity in the spectral and spatial dimensions, many works such as block-matching and 4-D filtering (BM4D) \cite{maggioni2012nonlocal} and the tensor dictionary learning \cite{peng2014decomposable} are also proposed.

These optimization-based HSI denoising methods are flexible to remove different types of noise \cite{he2015total,chen2017denoising} and can be even extended to tasks beyond denoising \cite{chang2020weighted,fu2017adaptive,he2019non,pengE3DTV}. However, their performance is significantly restricted by the matching degree of handcrafted prior and the underlying properties of HSI, which is difficult to ensure for complex real scenes. To address this problem, plug-and-play methods \cite{chan2016plug,danielyan2010image,zoran2011learning} integrate the optimization-based method with a learning-based prior, \ie, a plug-and-play Gaussian denoiser \cite{dabov2007image,zhang2018ffdnet}, to tackle complex noises that cannot be easily modeled by handcrafted prior. Specifically, Liu \emph{et al} \cite{liu2021hyperspectral} propose a fibered rank constrained tensor restoration framework and adopt BM3D \cite{dabov2007image} as an extra plug-and-play regularization.
In \cite{ma2020hyperspectral}, FFDNet \cite{zhang2018ffdnet} is used for local regularity, alongside the Kronecker-basis-representation-based tensor low-rankness for global structures regularity.

\subsection{Deep-Learning-based Methods}

Deep-learning-based methods have gained superior popularity in recent years. Inspired by 2D image denoising network DnCNN \cite{zhang2017beyond}, Chang \etal \cite{chang2018hsi} propose HSI-DeNet that learns multi-channel 2-D filters to model spectral correlation. Yuan \etal \cite{yuan2018hyperspectral} introduce residual network structure with a sliding window strategy for remote sensed HSI. To further exploit the spatial-spectral correlation, Dong \etal \cite{dong2019deep} designed a 3D U-net architecture. These methods have been successfully applied to different HSIs, but most of them are limited at exploring the inter-spectral correlations, which is significantly important for HSI denoising. To address such issue, Wei \etal \cite{wei20203} proposed a 3D convolutional quasi-recurrent neural network (QRNN3D), which combines the 3D convolution to extract spatial-spectral correlated features, and a quasi-recurrent network to integrate information from different bands in a global perspective. Based on it, Lai \etal \cite{lai2022deep} propose GRUNet that further improves the performance with residual blocks. Recent works also explore some hybrid approaches by borrowing key techniques from different types of methods. For example, T3SC \cite{bodrito2021trainable} is introduced as a hybrid method that combines sparse coding principles and deep neural networks. TRQ3DNet \cite{pang2022trq3dnet} augments the QRNN3D with a Uformer \cite{wang2022uformer} block to enhance the capabilities for capturing long-range spatial dependency. Compared with these methods, our method considers not only the inter-spectral correlation with a more powerful multi-head recurrent spectral attention but also the intra-spectral interaction with a progressive spectral channel attention block, which results in more discriminative features for the denoising. Moreover, we reexamine the low- and high-level feature fusion and propose an attentive skip connection for better feature aggregation, which is ignored in most existing works.

\section{Mixed Attention Network}

In this section, we first describe the overall architecture of our Mixed Attention Network (MAN). Then we provide detailed illustrations for each attention block, including multi-head recurrent spectral attention, progressive spectral channel attention, and attentive skip connection.

\vspace{-4mm}
\paragraph{Overall Architecture.}

\begin{figure}[t]
   \centering
   \includegraphics[width=1\linewidth]{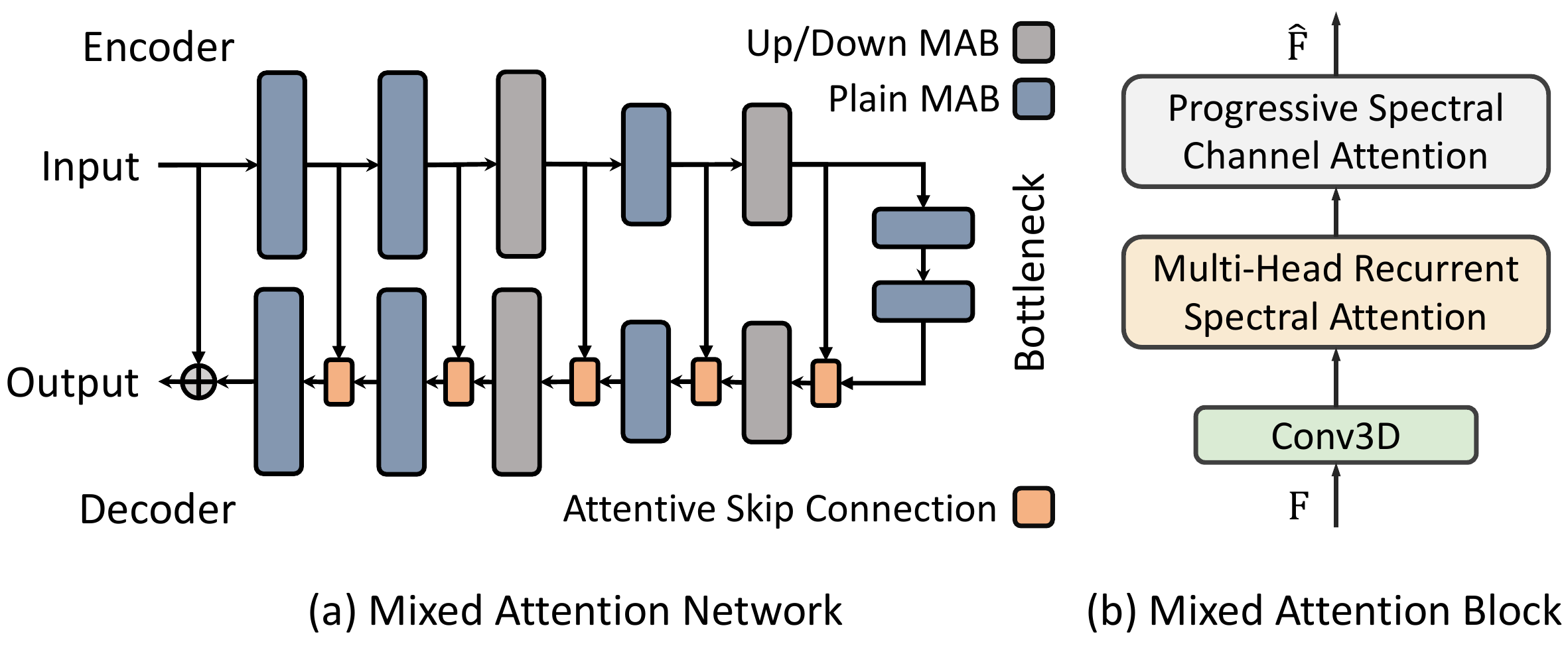}
   \caption{Overall architectures of the proposed mixed attention network and mixed attention block.}
   \label{fig:arch}
\end{figure}

The overall architecture of the proposed MAN is shown in Figure \ref{fig:arch}.
The input noisy image $\mathrm{X} \in \R^{H\times W\times S}$ is sequentially processed by a multi-level encoder to obtain multi-scale image features, which is then decoded by a symmetrically designed decoder to obtain the reconstructed clean image. We build the entire network by stacking Mixed Attention Blocks (MAB) that consist of a 3D convolution, a multi-head recurrent spectral attention, and a progressive spectral channel attention to explore the inter- and intra-spectral correlations. Each MAB takes input features $\mathrm{F} \in \R^{H\times W\times S\times C}$ and generates output features $\mathrm{\hat{F}} \in \R^{\hat{H}\times \hat{W}\times S\times \hat{C}}$ with the same spectral resolution $S$. Different from other encoder-decoder-based networks, MAN replaces the vanilla skip connection with an attentive one that adaptively controls which part of low-level features should flow forward from the encoder to the decoder.

\subsection{Multi-Head Recurrent Spectral Attention}

\begin{figure*}[t]
   \centering
   \includegraphics[width=1\linewidth]{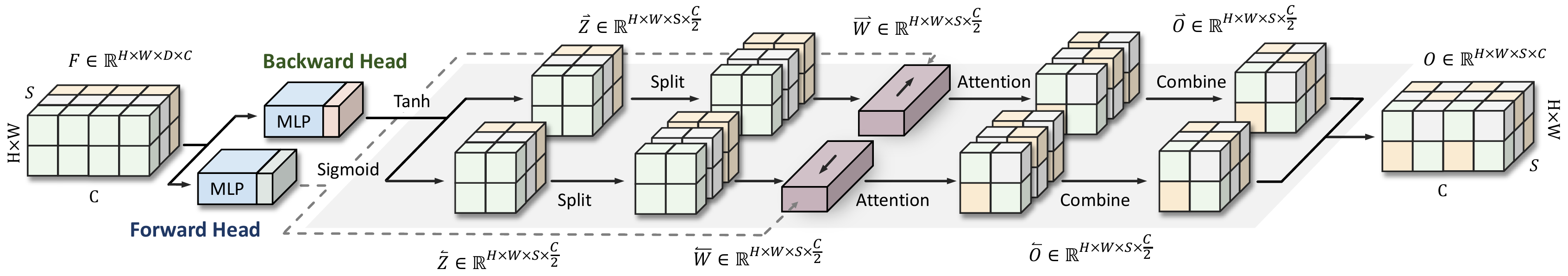}
   \caption{Illustration of the multi-head recurrent spectral attention. MLP stands for two-layer linear projections with Tanh activation. $C$, $S$, $HW$ denote the feature, spectral, and spatial dimensions, respectively.}
   \label{fig:mhrsa}
\end{figure*}

Different from RGB images, HSIs have a wide range of the spectrum, and the images across different bands share generally identical spatial content. Moreover, the pixel values with the same spatial location across different bands are correlated with specific spectral patterns, which are determined by the material of the object they belong to. This illustrates one important characteristic of HSI, \ie., inter-spectral correlation/similarity. Similar to non-local spatial correlation/similarity, it is straightforward to utilize inter-spectral correlations to perform denoising with methods like non-local means (NLM) \cite{buades2011non}. However, unlike spatial similar pixels, pixels along the spectrum have a different range of values, which makes naive NLM unsuitable as its averaging operation could break the spectral relationship of each pixel across the spectrum.

To address the above issue, we propose a Multi-Head Recurrent Spectral Attention (MHRSA) block that dynamically computes the weights for averaging the pixels along the spectrum for each band. Each band is assigned a different weight for averaging information from other bands,  which avoids the destruction of spectral dependency. An illustration of the MHRSA block is given in Figure \ref{fig:mhrsa}. Given a feature map $\mathrm{F} \in \R^{H\times W\times S\times C}$ extracted from the previous 3D convolution, it first uses two MLPs followed by two different activation functions to transform the input features into the candidate features $\mathrm{Z}$ and merging weights $\mathrm{W}$, as
\begin{equation}
   \begin{aligned}
      \mathrm{Z}                     & = \operatorname{tanh}(\operatorname{MLP_1}(\mathrm{F}))                   \\
      \mathrm{W}                     & = \operatorname{sigmoid}(\operatorname{MLP_2}(\mathrm{F}))                \\
      \operatorname{MLP}(\mathrm{X}) & = \mathrm{W_1} \cdot (\operatorname{tanh}(\mathrm{W_2} \cdot \mathrm{X}))
   \end{aligned}
\end{equation}
where $\mathrm{W_1}, \mathrm{W_2} \in \R^{C\times C}$. These processes can be equivalently considered as the query, key, and value projections in self-attention \cite{vaswani2017attention}. The one difference is that we directly compute the attention weight instead of obtaining the attention map through the covariance of key and query. The other is that we perform the attention operation through a recurrent merging step that only requires linear memory and time complexity other than the quadratic ones, which makes our method more suitable for high dimensional HSI data.

Specifically, the recurrent merging step for spectral mixing is performed by accumulating the candidate features $\mathrm{Z}$ for each band based on merging weight $\mathrm{W}$ as
\begin{equation}
   \mathrm{O_i} = (1-\mathrm{W_i}) \odot \mathrm{Z_i} + \mathrm{W_i} \odot \mathrm{O_{i-1}}
\end{equation}
where $\mathrm{O_i}$, $\mathrm{Z_i}$, $\mathrm{W_i}$ are the output features, candidate features, and merging weights of $i^{th}$ band, respectively. It can be observed that such a merging step fuse the features from all the previous bands $\mathrm{Z_i},i<j$ for $j^{th}$ band. Thus, it correlates the inter-spectral features and can potentially utilize the information from cleaner bands for denoising noisier bands.

Furthermore, though the above merging step progressively aggregates the inter-spectral features, each band can only see the features in one direction. It can lead to incomplete spectral context. To address the issue, we propose a multi-head attention in which we equally split the input features into two parts and perform parallel merging for each part with different directions. With such a strategy, we could achieve feature aggregation with global spectral context without any extra computation and parameters.

\subsection{Progressive Spectral Channel Attention}

\begin{figure}[t]
   \centering
   \includegraphics[width=1\linewidth]{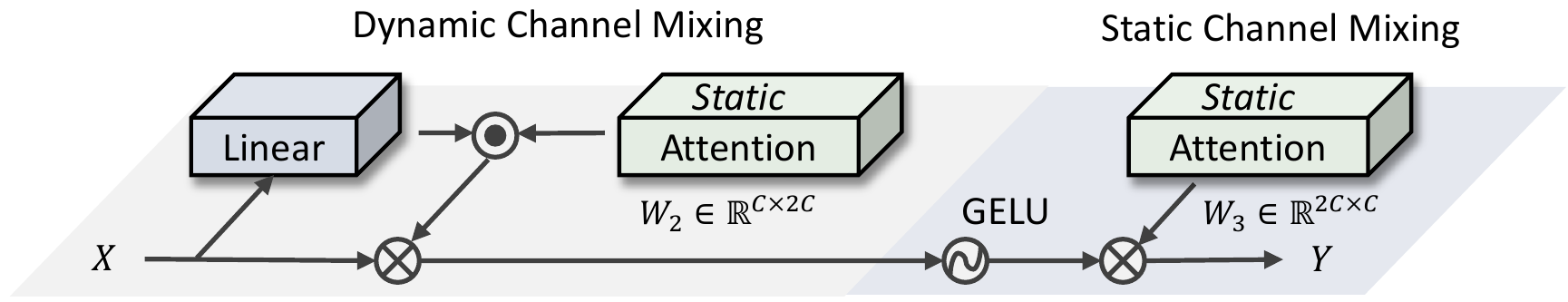}
   \caption{Illustration of progressive spectral channel attention. $\odot$, $\otimes$ denote element-wise- and original matrix multiplication. }
   \label{fig:ca}
\end{figure}

The MHRSA provides stronger capabilities for exploring the inter-spectral correlations, which makes our model powerful at borrowing the features from more informative bands for less informative ones. However, these advantages would be weakened if the features of each band itself are not discriminative enough. Therefore, we propose to further strengthen the features of each band by exploring the intra-spectral correlations with a Progressive Spectral Channel Attention (PSCA) module, as shown in Figure \ref{fig:ca}.

Our PSCA generalizes the band-wise Squeeze-Excitation (SE) \cite{hu2018squeeze} like channel attention to a pixel-wise operation with a progressive attention pipeline, which contains a Dynamic Channel Mixing (DCM) and a Static Channel Mixing (SCM) processes, as
\begin{equation}
   \mathrm{Y} = \operatorname{SCM}(\operatorname{GELU}(\operatorname{DCM}(X)))
\end{equation}
where $\mathrm{X}, \mathrm{Y}$ are the input and output feature maps.

\vspace{-4mm}
\paragraph{Static Channel Mixing.} Different from SE-like channel attention that performs channel-wise gating to control which channel should pass more information to subsequent layers, our SCM mix the information of all channels through a static attention map $\mathrm{W} \in \R^{C_{in}\times C_{out}}$ as,
\begin{equation}
   \operatorname{SCM}(\mathrm{X}) = \mathrm{X} \cdot \mathrm{W}.
\end{equation}
The attention map is jointly learned with the network training and it encodes the dataset level prior for selecting more diverse and important information across different features.

\vspace{-4mm}
\paragraph{Dynamic Channel Mixing.} To further strengthen the capability at identifying the essential features for each input HSI. We propose DCM that adopts a pixel-wise dynamic weight for channel mixing. Specifically, DCM first computes a scaling weight $\mathrm{S}$ for each input through a linear projection, $\mathrm{W_2} \in \R^{C\times C}$. Then, we rescale the static attention map $\mathrm{W_1}$ with the scaling weight through element-wise multiplication for each pixel location. Finally, we perform the channel mixing with the new attention map as
\begin{align}
   \operatorname{DCM}(\mathrm{X}) & = \mathrm{W_1}\cdot \mathrm{S} \odot \mathrm{X}, \\
   \mathrm{S}                     & = \mathrm{X} \cdot \mathrm{W_2}.
\end{align}
With these two channel mixing steps, we could obtain substantially better representations for each spectral band, thus could lead to better performance.

\subsection{Attentive Skip Connection}

Skip connection is the key part that distinguishes the U-Net \cite{dong2019deep} from other network architectures. It provides a direct but effective way to recover the low-level information that is lost during the downsampling operations of the commonly used encoder-decoder convolutional neural network.

The most commonly used additive skip connection is implemented as an addition between encoder and decoder features at the same depth.
Supposing features of encoder and decoder at the $i^{th}$ depth level are $\mathrm{F_e^{i}}$ and $\mathrm{F_d^{i}}$, respectively, the addition skip connection can be precisely described as
\begin{equation}
   \mathrm{F_d^{i+1}} = \mathrm{F_d^{i}} + \mathrm{F_e^{i}}.
\end{equation}
The major problem of it lies in the same weight on the shallow features from the encoder and highly processed features from the decoder. This could be problematic due to the imbalanced information density of different spectral bands and feature channels. For example, it would be more helpful to preserve more low-level features for cleaner spectral bands to retain the details while we might want more high-level features for noisy bands to properly remove all the noise.

\begin{figure}[t]
   \centering
   \includegraphics[width=1\linewidth]{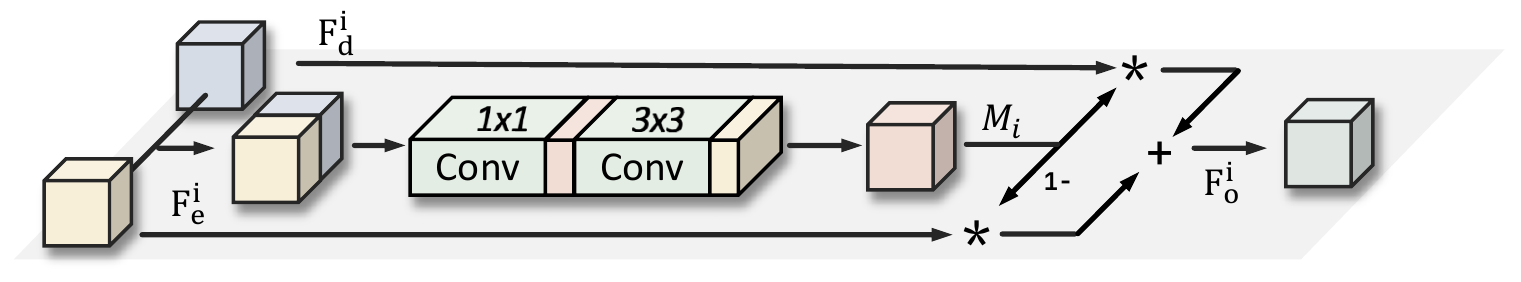}
   \caption{Illustration of the attentive skip connection block.}
   \label{fig:skip}
\end{figure}

To address the issue, we propose an Attentive Skip Connection (ASC) that explicitly weights the features from different sources using the attention weight computed from two convolution layers. By denoting the weight of $i^{th}$ depth level as $\mathrm{M_i}$, the detailed formulation is
\begin{equation}
   \begin{aligned}
      \mathrm{F}           & = \operatorname{LeakyReLU}(\operatorname{Conv1x1}([\mathrm{F_d^{i}}, \mathrm{F_e^{i}}])) \\
      \mathrm{M_i}         & =  \sigma(\operatorname{Conv3x3}(\mathrm{F})),                                           \\
      \mathrm{F_{d}^{i+1}} & = (1-\mathrm{M_i}) \odot \mathrm{F_d^{i}} + \mathrm{M_i} \odot \mathrm{F_e^{i}},
   \end{aligned}
\end{equation}
where $\sigma$ denotes the sigmoid function. A visualization of our attentive skip connection is shown in Figure \ref{fig:skip}. Our ASC adaptively strengthens the features of more important spatial locations and spectral bands through element-wise gating, thus leading to better reconstruction quality.

\vspace{-4mm}
\paragraph{Analysis and in-depth comparison.}

We provide an in-depth analysis of the difference between different types of skip connections.
Typically, the output of the skip connection module would be processed by the subsequent convolution layer. For additive skip connection, the computation can be formulated as,
\begin{equation}
   \mathrm{F_o} = \mathrm{W} \times (\mathrm{F_e} + \mathrm{F_d}) = \mathrm{W} \times \mathrm{F_e} + \mathrm{W} \times \mathrm{F_d}.
\end{equation}
where $\mathrm{F_o}$ is the output features from the subsequent convolution layer, $\mathrm{W}$ denotes the convolution kernel, $\times$ denotes convolution, and $\mathrm{F_e}$ and $\mathrm{F_d}$ are the features from the encoder and the decoder.  By using distributive law, we could know that we actually treat the features from the encoder and decoder as features in the same manifold, which are processed by the same convolution kernel.

For concat skip connection, the formulation would be,
\begin{equation}
   \mathrm{F_o} = \mathrm{W} \times [\mathrm{F_e}, \mathrm{F_d}] = \mathrm{W_1} \times \mathrm{F_e} + \mathrm{W_2} \times \mathrm{F_d}.
\end{equation}
where $\mathrm{W}$ is a convolution that transforms features with 2C channels to C channels. According to the properties of convolution, we could divide $\mathrm{W}$ into two parts, $\mathrm{W_1}$ and $\mathrm{W_2}$, which indicates that concat skip connection assigns different kernels for features from the encoder and decoder.

From the above analysis, we know that additive skip connection averages the features from the encoder and decoder extracted by the same convolution kernel. This might weaken the discriminative features due to the feature average. On the other hand, the concat skip connection extracts feature using two different kernels and then performs feature average. Thus, it is possible for it to generate features that avoid the problem of feature weakening in additive skip connection. However, its success requires two convolution kernels to simultaneously consider the feature extraction and reweighting, which is difficult to achieve. This not only includes the training difficulties but also the limitation of convolution itself for pixel-wise reweighting.

To address the aforementioned issues, our ASC separates the feature reweighting and extraction into two steps as,
\begin{equation}
   \begin{aligned}
      \mathrm{F_o} & =
      \mathrm{W} \times
      (\underbrace{\mathrm{M} \odot  \mathrm{F_e} + (1-\mathrm{M}) \odot \mathrm{F_d}}_{\text{feature reweighting}})
   \end{aligned}
\end{equation}
where $\mathrm{M}$ is the pixel-wise weight for reweighting. Such decomposition has two advantages. First, it avoids the need to learn two very discriminative convolutions as the concat skip connection, which makes the network concentrate more on finding more useful features. Second, it explicitly reweights the features from encoder and decoder, which distinguishes the more informative features to pass through the subsequent layers.

\section{Experiments}

In this section, we provide the experiments for simulated and real-world HSI denoising. An ablation study and discussion of each component are also included.

\input{tables/gaussian_v2.tex}

\input{figures/gaussian3.tex}

\subsection{Datasets}

We conducted experiments of simulated HSI denoising with ICVL \cite{arad2016sparse} and real-worlding HSI denoising with Urban and HSIDwRD \cite{zhang2021hyperspectral}. ICVL contains 201 clean HSI in the resolution of $1392 \times 1300$ that divides the spectrum into 31 spectral bands. Urban and HSIDwRD contain 1 HSI and 59 HSIs with 210 and 34 bands.  Following \cite{wei20203}, we split the ICVL into three parts, \ie, 100 images for training, 51 for validation, and 50 for testing. For training, we process the original HSIs into multiple overlapped smaller data cubes via even-stride cropping. The spatial resolution of each cube is $64 \times 64$ and the spectral resolution remains unchanged. Random rotation and scaling are also employed to further augment the dataset. For testing, the main region in the size of $512 \times 512 \times 31$ is used for ICVL, and 15 images in HSIDwRD are randomly selected. We use pretrained models of ICVL for Urban and HSIDwRD.

\input{tables/complex_v3.tex}
\input{figures/complex3.tex}

\subsection{Implementation Details}

Adam \cite{kingma2014adam} optimizer is adopted to minimize the mean square error. We follow the training strategy as \cite{wei20203}, with slight modifications on the setup of the learning rate. The strategy is briefly described here and we refer interested readers to \cite{wei20203} for more details. In short, the network is first trained on Gaussian noise with a fixed level of 50 for 20 epochs and then finetuned with a random noise level from a given set for 40 epochs to produce the first-stage Gaussian denoising model. In the second stage, the complex denoising model is obtained via another 30 epochs of fine-tuning using the trained Gaussian denoising model. The learning rate is set to $1\times 10^{-3}$ at first, but decayed by $0.1$ every 5 or 10 epochs, and finally reduced to $1\times 10^{-5}$ for each stage. The batch size is set to 16.

\subsection{Results on Gaussian Noise}

For the experiments with Gaussian noise, we evaluate our model with different noise strengths, including 30, 50, 70, and random strengths ranging from 30 to 70. The simulated noisy input is generated by adding zero-mean additive white Gaussian noise with a given variance. We use a single model to tackle different noise levels.

For systematical evaluation, we compare our method with four traditional HSI denoising methods, \ie, ITSReg \cite{xie2016multispectral}, KBR \cite{Qi2017Kronecker}, WLRTR \cite{chang2020weighted}, and NGmeet \cite{he2019non}, and five deep-learning ones, including HSID \cite{yuan2018hyperspectral}, QRNN3D \cite{wei20203}, T3SC \cite{bodrito2021trainable}, GRUNet \cite{lai2022deep}, and TRQ3D \cite{pang2022trq3dnet}. In the spirit of fairness, the hyperparameters in traditional methods are carefully tuned, and the deep-learning-based models are retrained if needed.

The quantitative and visual results are shown in Table \ref{fig:gaussian}. The comparisons of runtime and model size are provided in Table \ref{tab:speed}. It can be easily observed that our method achieves the best performance in all different settings with a large margin over the competing methods. Specifically, our MAN-L achieves over 0.6 dB PSNR improvement on average on four different settings while maintaining comparable parameters and runtime. our MAN-S achieves over 0.3 dB improvement against the best competing method, \ie, TRQ3D, with even fewer parameters. Moreover, our model is better at preserving spectral fidelity than previous methods, which is reflected by the improvement in SAM.

\input{tables/real.tex}

\subsection{Results on Complex Noise}

Unlike the simulated noisy HSI with Gaussian noise, the real HSI is usually corrupted by different types of complex noise, e.g., strip noise, impulse noise, and deadline noise. To evaluate the robustness of our method in real scenes, we also conduct experiments with different combinations of complex noise, following the settings in \cite{wei20203}.

We compare our method with eight recently developed methods including five deep-learning-based methods, \ie, HSID \cite{yuan2018hyperspectral}, QRNN3D \cite{wei20203}, GRUNet \cite{lai2022deep},  T3SC \cite{bodrito2021trainable}, and TRQ3D \cite{pang2022trq3dnet}, as well as four optimization-based ones, \ie, LRMR \cite{zhang2013hyperspectral}, LRTV \cite{he2015total}, NMoG \cite{chen2017denoising} and TDTV \cite{wang2017hyperspectral}. We choose a different set of optimization-based methods from the ones used for Gaussian noise because these methods only perform well for the noise settings they can solve.

Table \ref{fig:complex} gives the quantitative and visual comparison between our method against the competing ones. It can be seen that our method obtain substantial improvement on various metrics, especially on PSNR, which is over 1.5 dB on average, and ranged from 1.03 $\sim$ 1.7 dB.

\input{tables/ablation.tex}

\subsection{Results on Real-world Noise}

\paragraph{Urban.}
Figure \ref{fig:real} provide the visual result of our model on real-world noisy remotely sensed HSI, \ie, Urban. It can be seen that our model could properly remove the heavy noise while retaining the details. QRNN3D removes the most noise but loses the details such as stripes on the roof. HSID tends to produce blurry results than Ours.

\vspace{-3mm}
\paragraph{HSIDwRD.} We conduct experiments on the real natural HSI dataset \cite{zhang2021hyperspectral} with the first two leading optimization-based and CNN-based methods from complex denoising and the method proposed in \cite{zhang2021hyperspectral}. The quantitative results are shown in Table \ref{tab:real}. It can be observed that our results outperform the other ones in terms of all metrics, which demonstrates the effectiveness of our methods.

\setcounter{figure}{5}
\input{figures/urban2.tex}
\input{tables/speed}

\begin{figure}[t]
   \centering
   \includegraphics[width=0.95\linewidth]{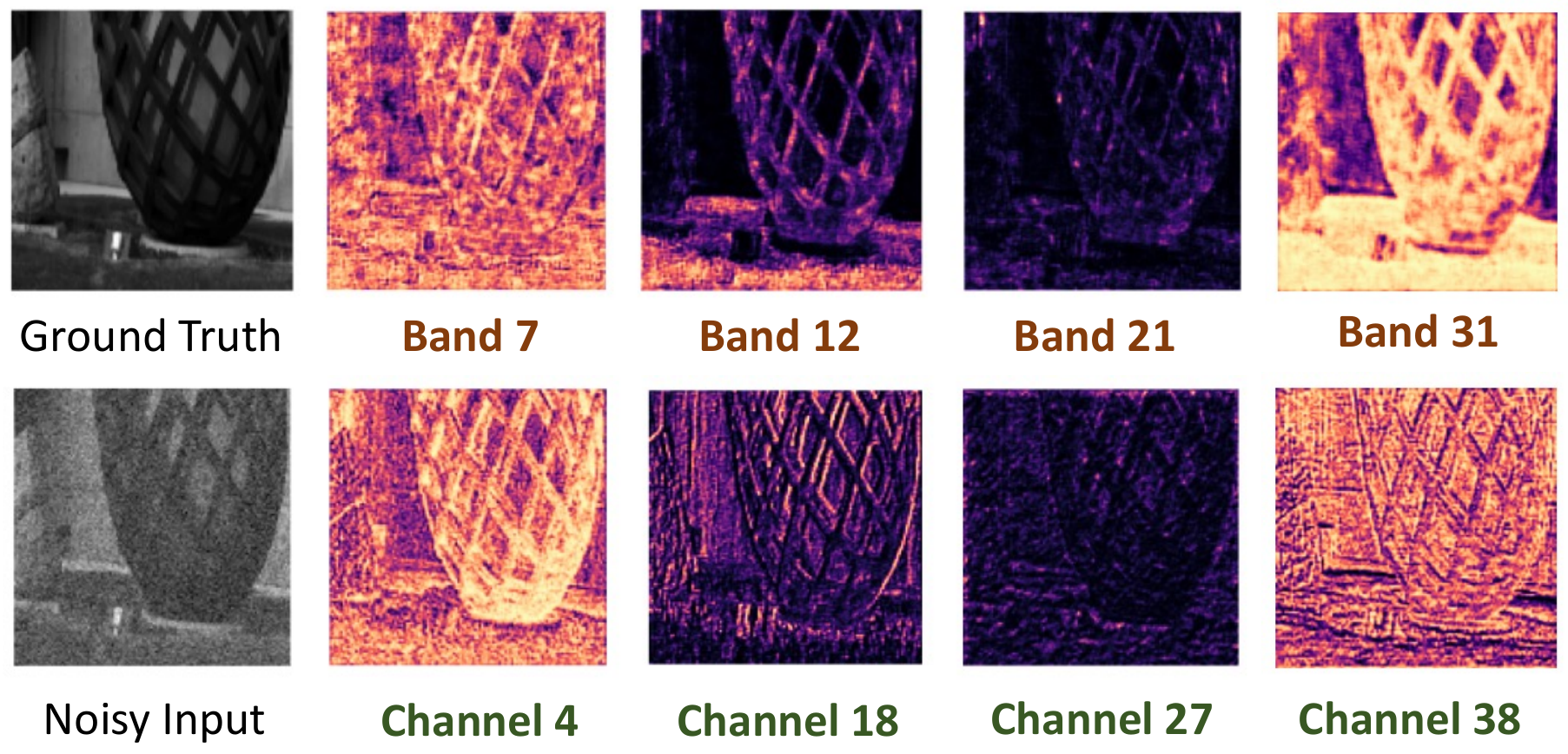}
   \vspace{-2mm}
   \caption{Visualization of the attention map of our attentive skip connection across different bands and channels.}
   \label{fig:ask}
\end{figure}

\begin{figure}[t]
   \centering
   \includegraphics[width=1\linewidth]{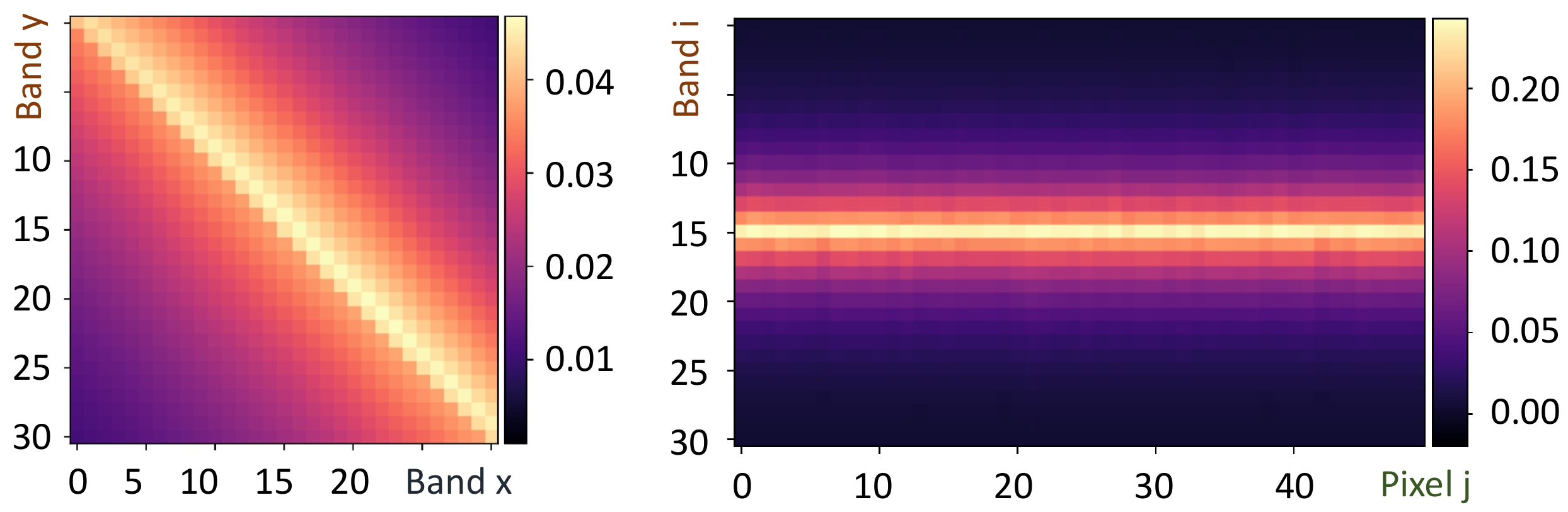}
   \vspace{-6mm}
   \caption{Visualization of the spectral attention map from our MHRSA. (a) The left part is the average attention map between different bands. (b) The right part is the attention distribution of $15^{th}$ band across different pixels.}
   \label{fig:sa-vis}
\end{figure}

\subsection{Discussions}

\paragraph{Ablation Study.}
To verify the effectiveness of each proposed component. We evaluate the PSNR improvement each component brings by adding them one by one, starting from a baseline model, \ie, UNet with 3D convolution and activation only. All the models are trained for the Gaussian denoising task and evaluated on 50 noise strength. The experimental results are shown in Table \ref{tab:ablation}. As we can observe, the introduction of MHRSA brings the most significant improvement. This indicates the importance of exploring inter-spectral correlations. The attentive skip connection contributes 0.35 dB improvement by taking into account low- and high-level features fusion. Our PSCA strengthens the features from MHRSA by considering intra-spectral relationships, which further provides 0.36 dB improvement.

\vspace{-3mm}
\paragraph{Visualization of Attentive Skip Connection.}

We visualize the attention map of the first attentive skip connection block for one sample. As is shown in Figure \ref{fig:ask}, the attention map varies across different bands and channels, which indicates that the equal attention of vanilla additive one might be less effective. In particular, it can be seen that the network learns to pay attention to different regions across different bands and channels, \ie, band12 and channel 21 for edges, and different sources, \eg, channel 18 for encoder, channel 38 for decoder.

\vspace{-3mm}
\paragraph{Visualization of Spectral Attention.}

Figure \ref{fig:sa-vis} shows the spectral attention map of one particular example HSI at the first MHRSA layer. It can be seen that our MHRSA pays attention to all bands with different weights, which is helpful for adaptively aggregating the useful features from different bands to assist the denoising. In general, the overall pattern is that each band pays more attention to the bands around it, but the attention distributions may vary with respect to pixels at different locations.

\section{Conclusion}

In this paper, we propose a mixed attention network for hyperspectral image denoising.
Our method introduces several key components to properly explore the inter- and intra-spectral correlations as well as the low- and high-level spatial-spectral feature interactions. Specifically, these are achieved with a multi-head recurrent spectral attention that recurrently merges the features across different bands, a progressive channel attention that progressively mixes the different features within each band, and an attentive skip connection that aggregates the features from encoder and decoder with different importance weight.  We perform extensive experiments on simulated and real-world noise, and it shows that our method outperforms the existing state-of-the-art methods with significant improvement while maintaining a smaller model size and running time.

\vspace{-3mm}
\paragraph{Ethical considerations and future work.} Our work has no ethical issues. In this work, we explore the proposed MHRSA with multi-head in two directions, but it is also possible to extend it to multi-axis in which we perform attention in different dimensions, \eg, channel and spectral.

%% file: tables/gaussian_v2.tex
\begin{table*}
\begin{center}
\small
\setlength{\tabcolsep}{0.1cm}
\begin{tabular}{@{}ccccccccccccccc@{}}
\toprule
\multirow{-2}{*}{Sigma} & \multirow{-2}{*}{Metric}                        & \multirow{-2}{*}{~Noisy~}    & \makecell{ITSReg \\ \cite{xie2016multispectral}} & \makecell{WLRTR \\\cite{chang2020weighted}}  & \makecell{~KBR~ \\ \cite{Qi2017Kronecker}}   & \makecell{NGmeet \\ \cite{he2019non}} & \makecell{HSID \\ \cite{yuan2018hyperspectral}}  & \makecell{QRNN3D \\ \cite{wei20203}} & \makecell{T3SC \\ \cite{bodrito2021trainable}}   & \makecell{GRUNet \\ \cite{lai2022deep}} & \makecell{TRQ3D \\\cite{pang2022trq3dnet}} & \makecell{MAN-S \\ (ours)} & \makecell{MAN-M \\ (ours)} & \makecell{MAN-L \\ (ours)}     \\ \midrule
\multirow{3}{*}{30}    & PSNR                                        & 18.59    & 41.48    & 42.62   & 41.48  & 42.99       & 41.72  & 42.22  & 42.36 & 42.84 & 43.25 & 43.83 & 44.07    & \textbf{44.17}    \\
                       & SSIM                                        & 0.110    & 0.961    & 0.988   & 0.984  & 0.989       & 0.987  & 0.988  & 0.986 & 0.989 & 0.990 & 0.991 & 0.991    & \textbf{0.991}        \\
                       & SAM                                         & 0.807    & 0.088    & 0.056   & 0.088  & 0.050       & 0.067  & 0.062  & 0.079 & 0.052 & 0.046 & 0.043 & 0.042    & \textbf{0.041}        \\ \midrule
\multirow{3}{*}{50}    & PSNR                                        & 14.15    & 38.88    & 39.72   & 39.16  & 40.26       & 39.39  & 40.15  & 40.47 & 40.75 & 41.30 & 41.60 & 41.84    & \textbf{41.94}    \\
                       & SSIM                                        & 0.046    & 0.941    & 0.978   & 0.974  & 0.980       & 0.980  & 0.982  & 0.980 & 0.983 & 0.985 & 0.985 & 0.986    & \textbf{0.986}        \\
                       & SAM                                         & 0.991    & 0.098    & 0.073   & 0.100  & 0.059       & 0.083  & 0.074  & 0.087 & 0.062 & 0.053 & 0.052 & 0.050    & \textbf{0.049}        \\  \midrule
\multirow{3}{*}{70}    & PSNR                                        & 11.23    & 36.71    & 37.52   & 36.71  & 38.66       & 37.77  & 38.30  & 39.05 & 39.02 & 39.86 & 40.05 & 40.32    & \textbf{40.40}    \\
                       & SSIM                                        & 0.025    & 0.923    & 0.967   & 0.961  & 0.974       & 0.972  & 0.974  & 0.974 & 0.977 & 0.980 & 0.980 & 0.981    & \textbf{0.981}        \\
                       & SAM                                         & 1.105    & 0.112    & 0.095   & 0.113  & 0.067       & 0.096  & 0.094  & 0.096 & 0.080 & 0.061 & 0.060 & 0.058    & \textbf{0.057}        \\ \midrule
\multirow{3}{*}{Blind} & PSNR                                        & 17.34    & 40.62    & 41.66   & 40.68  & 42.23       & 40.95  & 41.37  & 41.52 & 42.03 & 42.47 & 43.08 & 43.35    & \textbf{43.44}    \\
                       & SSIM                                        & 0.114    & 0.953    & 0.983   & 0.979  & 0.985       & 0.984  & 0.985  & 0.983 & 0.987 & 0.988 & 0.988 & 0.989    & \textbf{0.989}        \\
                       & SAM                                         & 0.859    & 0.087    & 0.064   & 0.080  & 0.053       & 0.072  & 0.068  & 0.085 & 0.057 & 0.054 & 0.046 & 0.045    & \textbf{0.044}        \\            
                       \bottomrule
\end{tabular}
\end{center}
\vspace{-5mm}
\label{tab:denoise-gaussian}
\end{table*}

%% file: figures/gaussian3.tex
\begin{figure*}[h]
    \begin{center}
      \begin{subfigure}[b]{0.095\textwidth}
          \centering
          \includegraphics[width=\textwidth]{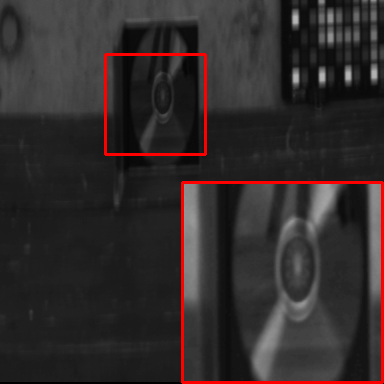}
          \caption{GT}
      \end{subfigure}
     \begin{subfigure}[b]{0.095\textwidth}
        \centering
        \includegraphics[width=\textwidth]{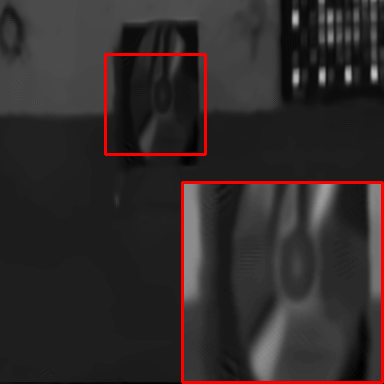}
        \caption{WLRTR}
    \end{subfigure}
      \begin{subfigure}[b]{0.095\textwidth}
          \centering
          \includegraphics[width=\textwidth]{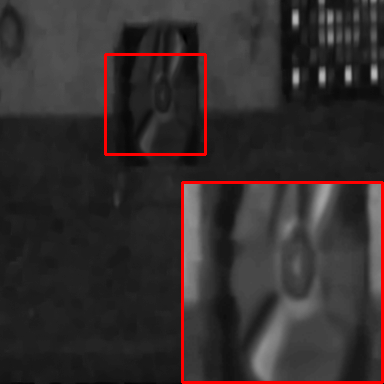}
          \caption{KBR}
      \end{subfigure}
      \begin{subfigure}[b]{0.095\textwidth}
          \centering
          \includegraphics[width=\textwidth]{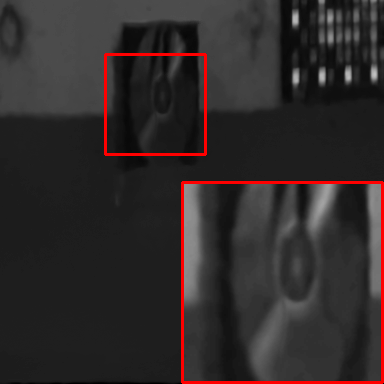}
          \caption{NGmeet}
      \end{subfigure}
      \begin{subfigure}[b]{0.095\textwidth}
          \centering
          \includegraphics[width=\textwidth]{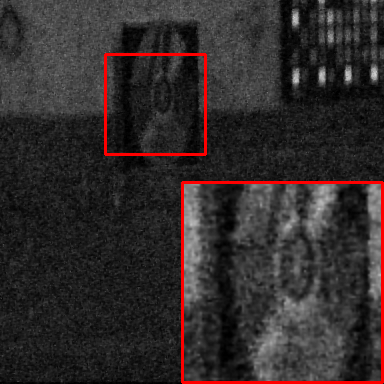}
          \caption{HSID}
      \end{subfigure}
     \begin{subfigure}[b]{0.095\textwidth}
         \centering
         \includegraphics[width=\textwidth]{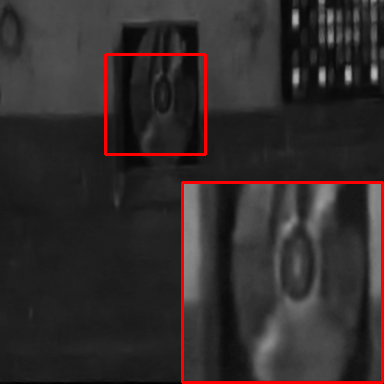}
         \caption{QRNN3D}
     \end{subfigure}
     \begin{subfigure}[b]{0.095\textwidth}
         \centering
         \includegraphics[width=\textwidth]{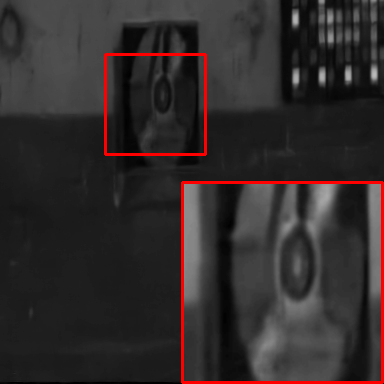}
         \caption{T3SC}
     \end{subfigure}
     \begin{subfigure}[b]{0.095\textwidth}
         \centering
         \includegraphics[width=\textwidth]{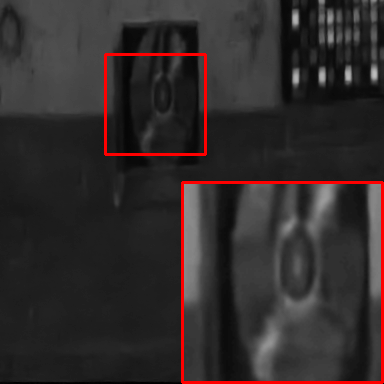}
         \caption{GRUNet}
     \end{subfigure}
      \begin{subfigure}[b]{0.095\textwidth}
         \centering
         \includegraphics[width=\textwidth]{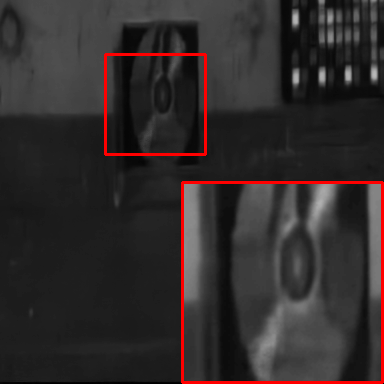}
         \caption{TRQ3D}
     \end{subfigure}
     \begin{subfigure}[b]{0.095\textwidth}
         \centering
         \includegraphics[width=\textwidth]{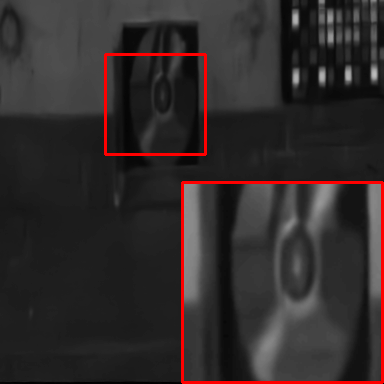}
         \caption{MAN-M}
     \end{subfigure}
	\end{center}
	\vspace{-5mm}
    \captionof{table}{Simulated Gaussian noise removal results under several noise levels on ICVL. \emph{Blind} suggests each image is corrupted by Gaussian noise with random sigma (ranged from 30 to 70). Visual results show the $20^{th}$ band under noise level 50.}
    \label{fig:gaussian}
    \vspace{-2mm}
 \end{figure*}

%% file: tables/complex_v3.tex
\begin{table*}
\begin{center}
\setlength{\tabcolsep}{0.137cm}
\small
\begin{tabular}{@{}cccccccccccccc@{}}
\toprule
\multirow{-2}{*}{Type} & \multirow{-2}{*}{Metric}                        & \multirow{-2}{*}{~Noisy~}  & \makecell{LRMR \\ \cite{zhang2013hyperspectral}}   & \makecell{LRTV \\ \cite{he2015total}} & \makecell{NMoG \\\cite{chen2017denoising}}  & \makecell{TDTV \\ \cite{wang2017hyperspectral}}   & \makecell{HSID \\ \cite{yuan2018hyperspectral}} & \makecell{QRNN3D \\ \cite{wei20203}}  & \makecell{T3SC \\ \cite{bodrito2021trainable}} & \makecell{GRUNet \\ \cite{lai2022deep}}  & \makecell{TRQ3D \\ \cite{pang2022trq3dnet}} & \makecell{MAN-S \\ (ours)} & \makecell{MAN-M \\ (ours)}     \\ \midrule
\multirow{3}{*}{G+Stripe}    & PSNR                                  & 17.80   &32.62    & 33.49    & 33.87   & 37.67  & 37.77       & 42.35  & 40.85  & 42.39 & 43.05 & 44.24 & \textbf{44.60}  \\
                       & SSIM                                        & 0.159   &0.717    & 0.905    & 0.799   & 0.940  & 0.942       & 0.976  & 0.986  & 0.991 & 0.992 & 0.993 & \textbf{0.994}  \\
                       & SAM                                         & 0.910   &0.187    & 0.078    & 0.265   & 0.081  & 0.104       & 0.055  & 0.072  & 0.050 & 0.043 & 0.039 & \textbf{0.038}  \\ \midrule
\multirow{3}{*}{G+Deadline}    & PSNR                                & 17.61   &31.83    & 32.37    & 32.87   & 36.15  & 37.65       & 42.23  & 39.54  & 42.11 & 42.95 & 44.13 & \textbf{44.51}  \\
                       & SSIM                                        & 0.155   &0.709    & 0.895    & 0.797   & 0.930  & 0.940       & 0.976  & 0.983  & 0.991 & 0.992 & 0.993 & \textbf{0.994}  \\
                       & SAM                                         & 0.917   &0.227    & 0.115    & 0.276   & 0.099  & 0.102       & 0.056  & 0.096  & 0.050 & 0.044 & 0.040 & \textbf{0.038}        \\ \midrule
\multirow{3}{*}{G+Impulse}    & PSNR                                 & 14.80   &29.70    & 31.56    & 28.60   & 36.67  & 35.00       & 39.23  & 36.06  & 40.70 & 41.27 & 41.88 & \textbf{41.97} \\
                       & SSIM                                        & 0.114   &0.623    & 0.871    & 0.652   & 0.935  & 0.899       & 0.945  & 0.952  & \textbf{0.985} & 0.983 & 0.982 & 0.979 \\
                       & SAM                                         & 0.926   &0.311    & 0.242    & 0.486   & 0.094  & 0.174       & 0.109  & 0.203  & \textbf{0.067} & 0.075 & 0.090 & 0.095  \\  \midrule
\multirow{3}{*}{G+Mixture}    & PSNR                                 & 14.08   &28.68    & 30.47    & 27.31   & 34.77  & 34.05       & 38.25  & 34.48  & 38.51 & 40.27 & 41.03 & \textbf{41.18} \\
                       & SSIM                                        & 0.099   &0.608    & 0.858    & 0.632   & 0.919  & 0.888       & 0.938  & 0.946  & 0.981 & 0.983 & \textbf{0.981} & 0.979         \\
                       & SAM                                         & 0.944   &0.353    & 0.287    & 0.513   & 0.113  & 0.181       & 0.107  & 0.228  & \textbf{0.081} & 0.075 & 0.092 & 0.097      \\
                       \bottomrule
\end{tabular}
\end{center}
\vspace{-5mm}
\label{tab:denoise-complex}
\end{table*}

%% file: figures/complex3.tex
\begin{figure*}[h]
    \begin{center}
      \begin{subfigure}[b]{0.095\textwidth}
          \centering
          \includegraphics[width=\textwidth]{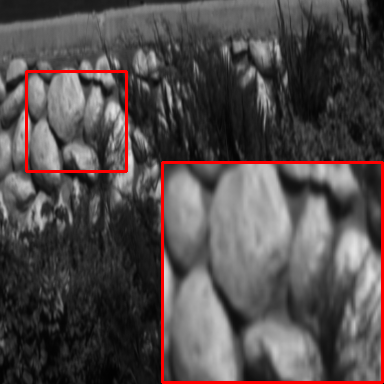}
          \caption{GT}
      \end{subfigure}
     \begin{subfigure}[b]{0.095\textwidth}
        \centering
        \includegraphics[width=\textwidth]{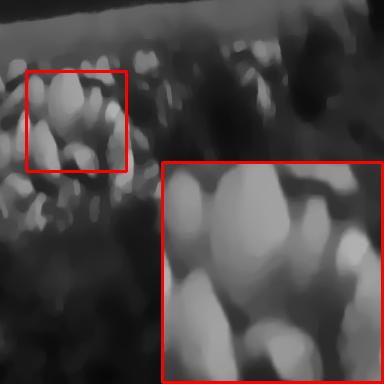}
        \caption{LRTV}
    \end{subfigure}
      \begin{subfigure}[b]{0.095\textwidth}
          \centering
          \includegraphics[width=\textwidth]{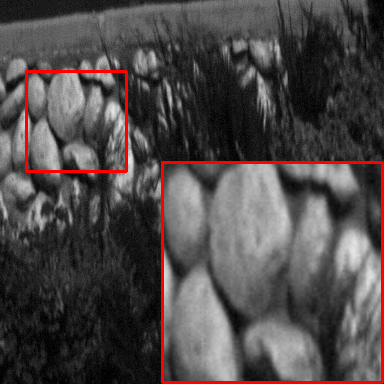}
          \caption{NMoG}
      \end{subfigure}
      \begin{subfigure}[b]{0.095\textwidth}
          \centering
          \includegraphics[width=\textwidth]{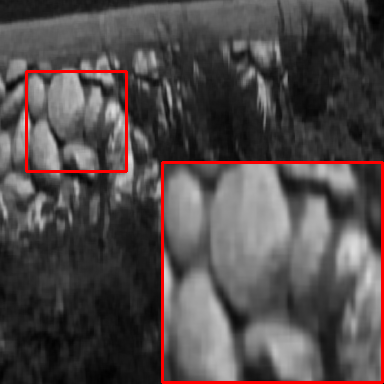}
          \caption{TDTV}
      \end{subfigure}
      \begin{subfigure}[b]{0.095\textwidth}
          \centering
          \includegraphics[width=\textwidth]{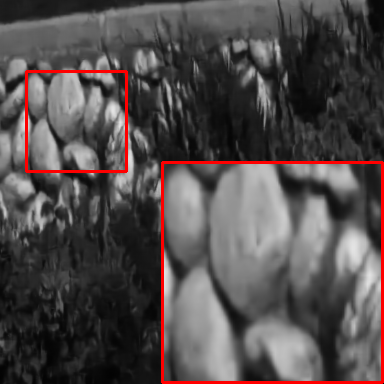}
          \caption{HSID}
      \end{subfigure}
     \begin{subfigure}[b]{0.095\textwidth}
         \centering
         \includegraphics[width=\textwidth]{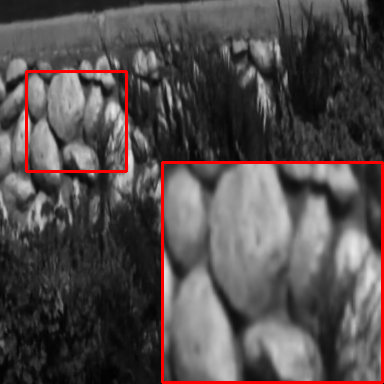}
         \caption{QRNN3D}
     \end{subfigure}
     \begin{subfigure}[b]{0.095\textwidth}
         \centering
         \includegraphics[width=\textwidth]{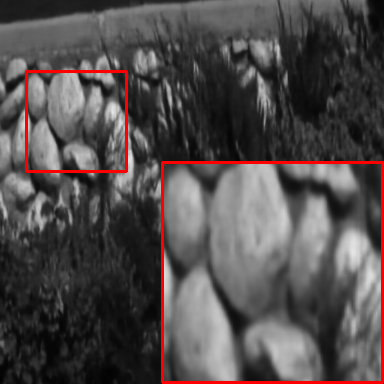}
         \caption{T3SC}
     \end{subfigure}
     \begin{subfigure}[b]{0.095\textwidth}
         \centering
         \includegraphics[width=\textwidth]{imgs/complex-stripe/grunet.png}
         \caption{GRUNet}
     \end{subfigure}
      \begin{subfigure}[b]{0.095\textwidth}
         \centering
         \includegraphics[width=\textwidth]{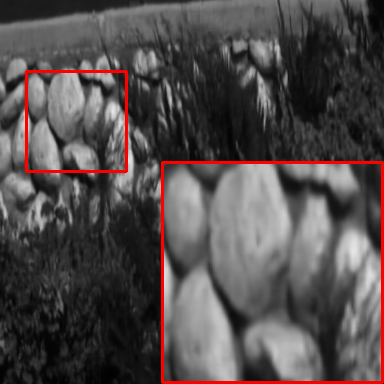}
         \caption{TRQ3D}
     \end{subfigure}
     \begin{subfigure}[b]{0.095\textwidth}
         \centering
         \includegraphics[width=\textwidth]{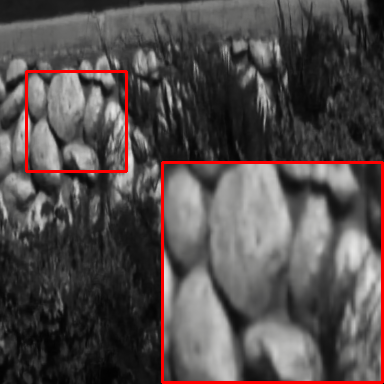}
         \caption{MAN-M}
     \end{subfigure}
	\end{center}
	\vspace{-5mm}
    \captionof{table}{Simulated complex noise removal results under several noise types on ICVL. Visual results show the $20^{th}$ band under stripe noise.}
    \label{fig:complex}
    \vspace{-2mm}
 \end{figure*}

%% file: tables/real.tex
\begin{table}[t]
   \small
   \centering
   \setlength{\tabcolsep}{0.077cm}
   \begin{tabular}{@{}ccccccc@{}}
      \toprule
\makecell{Metrics\\~}           & \makecell{NMoG\\ \cite{chen2017denoising}} & \makecell{TDTV \\ \cite{wang2017hyperspectral}}   & \makecell{QRNN3D \\ \cite{wei20203}} & \makecell{HSID \\  \cite{yuan2018hyperspectral}} & \makecell{HSIDwRD \\ \cite{zhang2021hyperspectral}} & \makecell{MAN-M\\ (ours)}  \\ \midrule
        PSNR        & 30.90    &31.14 & 31.13 & 31.05  & 31.23   &\textbf{31.36}  \\
       SSIM         & 0.907    &0.881 & 0.940 & 0.934  & 0.939   &\textbf{0.940} \\
       SAM          & 1.761    &1.853 & 0.094 & 0.096  & 0.092   &\textbf{0.092} \\
      \bottomrule
   \end{tabular}
   \vspace{-2mm}
   \caption{Quantitative results on HSIDwRD \cite{zhang2021hyperspectral}.}
   \label{tab:real}
   \vspace{-2mm}
\end{table}

%% file: tables/ablation.tex
\begin{table}[t]
   \small
   \centering
   \setlength{\tabcolsep}{0.16cm}
   \begin{tabular}{@{}ccccccc@{}}
      \toprule
        MHRSA        & ASC         & PSCA          & Params & Runtime & PSNR&SAM  \\ \midrule
        -          &     -       &    -          & 0.43  & 0.25   &39.98 & 0.072 \\
       \checkmark &     -       &     -         & 0.50  & 0.46   &41.13 & 0.058 \\
       \checkmark & \checkmark &  -            & 0.81  & 0.55   &41.48 & 0.053 \\
       \checkmark &  \checkmark          & \checkmark   & \textbf{0.89}  & \textbf{0.65}   & \textbf{41.84} & \textbf{0.050} \\
      \bottomrule
   \end{tabular}
   \vspace{-2mm}
   \caption{Ablation study of the proposed network modules.}
   \label{tab:ablation}
\end{table}

%% file: figures/urban2.tex
\begin{figure*}[h]
     \begin{center}
     \begin{subfigure}[b]{0.135\textwidth}
         \centering
         \includegraphics[width=\textwidth]{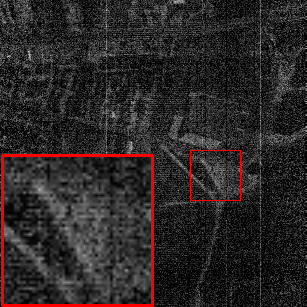}
         \caption{Noisy}
     \end{subfigure}
     \hfill
     \begin{subfigure}[b]{0.135\textwidth}
        \centering
        \includegraphics[width=\textwidth]{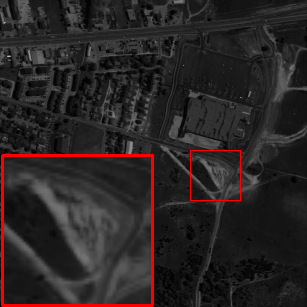}
        \caption{NMoG \cite{chen2017denoising}}
    \end{subfigure}
    \hfill
     \begin{subfigure}[b]{0.135\textwidth}
         \centering
         \includegraphics[width=\textwidth]{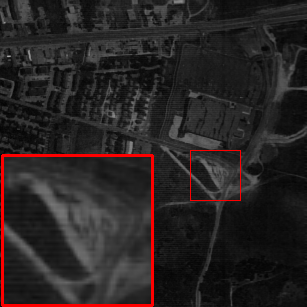}
         \caption{TDTV \cite{wang2017hyperspectral}}
     \end{subfigure}
      \hfill
     \begin{subfigure}[b]{0.135\textwidth}
         \centering
         \includegraphics[width=\textwidth]{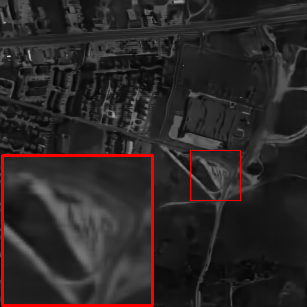}
         \caption{HSID \cite{yuan2018hyperspectral}}
     \end{subfigure}
      \hfill
     \begin{subfigure}[b]{0.135\textwidth}
         \centering
         \includegraphics[width=\textwidth]{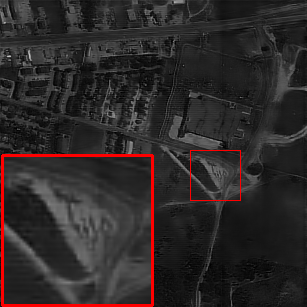}
         \caption{QRNN3D \cite{wei20203}}
     \end{subfigure}
     \hfill
   	 \begin{subfigure}[b]{0.135\textwidth}
         \centering
         \includegraphics[width=\textwidth]{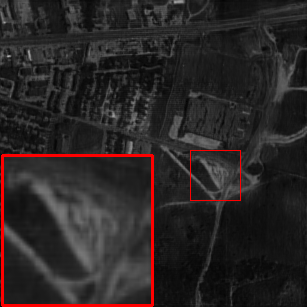}
         \caption{GRUNet \cite{lai2022deep}}
     \end{subfigure}
      \hfill
     \begin{subfigure}[b]{0.135\textwidth}
         \centering
         \includegraphics[width=\textwidth]{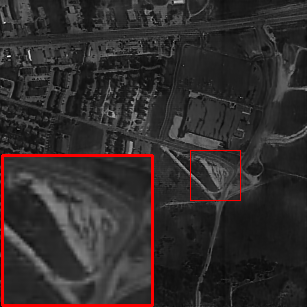}
         \caption{MAN(Ours)}
     \end{subfigure}
     \end{center}
     \vspace{-5mm}
   \caption{Real world noise removal results at $207^{th}$ band of Urban Dataset.}
   \label{fig:real}
   \vspace{-2mm}
\end{figure*}

%% file: tables/speed.tex
\begin{table*}
\begin{center}

\setlength{\tabcolsep}{0.14cm}
\small
\begin{tabular}{@{}lcccccccccccc@{}}
\toprule
 \multirow{-2}{*}{~Method}  
  & \makecell{ITSReg \\ \cite{xie2016multispectral}} & \makecell{WLRTR \\\cite{chang2020weighted}}  
  & \makecell{~KBR~ \\ \cite{Qi2017Kronecker}}   & \makecell{NGmeet \\ \cite{he2019non}} 
  & \makecell{HSID \\ \cite{yuan2018hyperspectral}}   & \makecell{QRNN3D \\ \cite{wei20203}} 
  & \makecell{T3SC \\ \cite{bodrito2021trainable}}   & \makecell{GRUNet \\ \cite{lai2022deep}} & \makecell{TRQ3D \\ \cite{pang2022trq3dnet}} 
 & \makecell{MAN-S \\ (ours)} & \makecell{MAN-M \\ (ours)} & \makecell{MAN-L \\ (ours)}     
\\ \midrule
~PSNR  & 38.88  & 39.16  & 38.70       & 40.26  & 39.39 & 40.15 & 40.47 & 40.75  & 41.30 & \textbf{41.60}  & \textbf{41.84}  & \textbf{41.94}  \\ 
~Params (M)    & -   & -    & -   & -  & 0.40       & 0.86    & 0.83     & 14.2  & 0.68  & \textbf{0.50}  & \textbf{0.89}  & \textbf{1.39}    \\
~Runtime (s)   &  907   & 1600     & 1755   & 166    & 0.48   & 0.44 & 0.95  & 0.87       & 0.47    &  \textbf{0.49} & \textbf{0.65} & \textbf{0.97}      \\            
                       \bottomrule
\end{tabular}
\end{center}
\vspace{-5mm}
\caption{Performance-Params-Runtime comparisons with the SOTA methods. Evaluated with Gaussian denoising (sigma=50) on ICVL.}
\label{tab:speed}
\end{table*}